# Edge Intelligence in Civil Aviation: Paradigms, Techniques, and Applications

Wenbin Li*
*China Southern Digital Intelligence Technology (Guangdong) Co., Ltd.*
*China Southern Airlines Co., Ltd.*
Guangzhou, China
li_wenbin@csair.com

Zhongtian Liao
*China Southern Digital Intelligence Technology (Guangdong) Co., Ltd.*
*China Southern Airlines Co., Ltd.*
Guangzhou, China
liaozhongtian@csair.com

Bolin Liu
*China Southern Digital Intelligence Technology (Guangdong) Co., Ltd.*
*China Southern Airlines Co., Ltd.*
Guangzhou, China
liu_bolin@csair.com

Yongjie Zhou
*China Southern Digital Intelligence Technology (Guangdong) Co., Ltd.*
*China Southern Airlines Co., Ltd.*
Guangzhou, China
zhouyongjie@csair.com

Jingling Wu
*China Southern Digital Intelligence Technology (Guangdong) Co., Ltd.*
*China Southern Airlines Co., Ltd.*
Guangzhou, China
wujingl@csair.com

Xiaoyong Lin
*China Southern Digital Intelligence Technology (Guangdong) Co., Ltd.*
*China Southern Airlines Co., Ltd.*
Guangzhou, China
linxy@csair.com

Jing Chen
*China Southern Digital Intelligence Technology (Guangdong) Co., Ltd.*
*China Southern Airlines Co., Ltd.*
Guangzhou, China
chenjingc@csair.com

***Abstract*—Civil aviation is safety critical and its operations, from flight decks and towers to ramps and maintenance, generate massive, heterogeneous data at the network edge. Yet cloud centric deployment of large Artificial Intelligence (AI) models often produces high task latency, lacks offline capability in communication denied environments, and requires centralizing sensitive data, raising privacy and sovereignty risks. Edge AI moves perception, prediction, and decision logic closer to the data producers via compression, collaborative inference, and split learning, thereby reducing latency, bandwidth, and exposure while enabling graceful operation during disconnections. This paper provides a panoramic view and a common understanding of edge intelligence tailored to civil aviation. We firstly articulate the operational motivations for edge AI, and then review recent techniques for edge inference and edge learning. We then introduce the organizational computing paradigms and the respective configurations in civil aviation environments; finally, we describe the emerging applications and the future research trends of edge intelligence in civil aviation. We argue that a refined edge solution can complement cloud foundations to deliver low‑latency, privacy‑preserving, and resilient AI services across the civil aviation lifecycle.**

## I. Introduction

The civil aviation industry is undergoing a digital transformation as airlines, airports, and air traffic management adopt AI and data-driven technologies to enhance safety, efficiency, and passenger experience [1]. Nowadays, the civil aviation ecosystem increasingly relies on digital sensors, IP cameras, software-defined radios, avionics buses and maintenance information systems that produce continuous multimodal streams. AI techniques can mine such data to detect safety risks and suggest preventive actions, as well as optimize routes and fuel usage in real time [2]. Embracing AI promises improved operational resilience and cost savings in this highly competitive sector.

Currently, most of aviation AI solutions rely on large deep learning models hosted in cloud data centers. While cloud-based solutions leverage powerful computing clusters, it also incurs drawbacks that are particularly problematic in aviation contexts. Firstly, sending data (e.g., from an aircraft or airport sensor) to a remote cloud and awaiting inference results may not meet the low latency requirements of flight operations or safety. Secondly, connectivity constraints further complicate this, as aircraft in flight or remote airports cannot always maintain high-bandwidth, low-latency links to cloud servers, and correspondingly the capability of an AI solution to work offline is very limited; furthermore, another critical concern is data privacy and security. Aviation data (e.g., aircraft sensor readings, maintenance logs, or passenger information) are often sensitive, and transmitting large volumes of raw data to external clouds raises risks of leakage and regulatory [3]. In such context, conventional cloud-based AI solutions can suffer from high end-to-end latency, reliance on continuous connectivity, and potential privacy breaches. Such limitations motivate a shift toward more localized intelligence in aviation.

Edge intelligence refers to embedding AI capabilities at the network edge, such as end devices, embedded processors, and local edge servers, rather than exclusively in centralized clouds [4]. This paradigm combines edge computing (i.e., processing data near its source) with artificial intelligence, enabling models to be trained and run on distributed edge nodes [3]. By pushing computation closer to where aviation data is generated (e.g., aircraft onboard systems, airport equipment, control towers, etc.), edge AI can drastically reduce communication latency and improve responsiveness. Importantly, it also keeps data local, alleviating privacy and bandwidth issues by avoiding bulk data uploads. The concept of edge intelligence has rapidly evolved with advances in Internet of Things (IoT) hardware and on-device AI optimization. The work in [5] formalized edge intelligence as the convergence of edge computing and AI, describing levels from cloud-edge co-inference down to fully on-device AI. Such foundational works highlight that edge intelligence naturally addresses the shortcomings of cloud-centralized AI by exploiting distributed resources. Recent surveys further emphasize that edge AI is a cutting-edge direction in AI, enabling real-time and reliable services by utilizing nearby computing nodes [6]. In the aviation domain, edge intelligence aligns well with the industry's needs for safety-critical reliability, real-time operation, and data sovereignty. Smart airport systems [7] now deploy edge AI for video analytics in surveillance and runway monitoring on-site, rather than relying on remote data center. Airlines are investing in edge cloud systems aboard aircraft to preprocess telemetry and even run certain inference tasks mid-flight, which can save costly minutes in decision-making, estimated at $600 per minute of delay [8]. Overall, edge AI brings computation and decision-making on-premises into the cockpit, the airport, and the airspace: a paradigm shifting highly congruent with aviation's stringent performance and security requirements.

This paper provides a unified technical background and a comprehensive review of edge intelligence as applied to civil aviation, with the following contributions:

- We review the core edge AI techniques that enable deep learning models to run efficiently on edge devices.
- We define and discuss four key organizational computing paradigms for distributing intelligence between cloud, edge, and devices in aviation systems.
- We examine current and emerging applications in civil aviation where edge intelligence is being deployed.
- We discuss future trends and visions, such as the development of lightweight multimodal models for edge use, the rise of edge-capable large language models, and the co-optimization of edge and cloud in hybrid architectures.

The remainder of this paper is structured as follows. Section 2 introduces the representative edge AI techniques divided into edge inference optimization and edge learning methods, and discusses the advantages of edge deployment in terms of latency, privacy, bandwidth, and so on. Section 3 describes four paradigms for organizing computation between cloud and edge in aviation systems. Section 4 presents several edge intelligence application areas in civil aviation. Section 5 explores future trends and visions for edge AI in aviation. Section 6 concludes the paper.

## II. Edge Intelligence Techniques

Edge intelligence in aviation is enabled by a suite of techniques for edge inference and edge learning.

### *A. Edge inference: quantization, pruning, distillation and complements*

- **Quantization.** Quantization reduces the numerical precision of model parameters and computations. The Phi-3-mini (3.8B) [9] is trained with 3.3T high-quality tokens and distilled alignment data, reaching 69% on MMLU and 8.38 on MT-bench, achieving performance comparable to GPT-3.5 and supporting efficient 4-bit deployment (>12 tokens/s on iPhone 14 with only 1.8 GB memory usage). SmoothQuant enables training free post training quantization (PTQ) to W8A8 for billion scale LLMs by redistributing activation ranges, achieving near lossless accuracy even on 530B parameter models [10]. QLoRA back propagates through a frozen 4-bit model into low rank adapters, enabling single GPU finetuning of 65B LLMs without degradation [11]. AWQ identifies a small set of "salient" weight channels via activation statistics and scales them to preserve accuracy under 3 to 4-bit weight only quantization, demonstrating strong generalization to instruction tuned and multi modal models [12][13]. SqueezeLLM further combines dense and sparse decomposition to push ultra-low precision while retaining accuracy and speedup [14].
- **Pruning.** Pruning techniques eliminate redundant weights or neurons from a neural network. SparseGPT shows one shot pruning of OPT 175B and BLOOM 176B to 50–60% sparsity with minimal perplexity increase, making large models lighter without retraining [15]. Wanda prunes by the product of weight magnitudes and input activation norms, outperforming magnitude pruning and competing with more expensive methods; its 2024 ICLR version provides extensive LLaMA 2 results [16]. Recent structured variants explore block wise gradients and adaptive structure search to produce deployment friendly sparsity patterns [17]. The work in [18] proposed Llama-3.1-Minitron-4B, compressing Llama-3.1-8B by width-based pruning and knowledge distillation, which reduced parameters from 8B to 4.5B, achieved comparable accuracy to the teacher model, and delivered up to ~1.8× higher inference throughput.
- **Knowledge distillation.** Knowledge distillation transfers the knowledge from a large "teacher" model (or ensemble of models) into a smaller "student" model. The student is trained to mimic the output distributions (soft predictions) of the teacher, thereby achieving similar accuracy with far fewer parameters. For speech, Distil Whisper compresses Whisper into students that are ~6× faster and ~49% smaller with ≤1% WER loss by selecting high quality pseudo labels at scale [19]. For LLMs, surveys systematize algorithmic choices (white vs. black box KD, data augmentation, skill oriented distillation) and report competitive accuracy for compact students on reasoning and instruction following [20][21]. DeepSeek-R1-Distill-Qwen-7B, distilled from DeepSeek-R1, with Qwen2.5-Math-7B as the student model, preserves the 7B parameter size while boosting performance, achieving a pass@1 score of 92.8% on MATH-500 [22]. Lightweight adapters such as LLaMA Adapter efficiently inject task conditioning with only ~1.2M trainable parameters on LLaMA 7B [23].

- **Complementary accelerants.** Early exit mechanisms and split computing decide at runtime whether to finish inference locally, continue on an edge server, or escalate to cloud to meet latency or energy budgets [24]. Collaborative DNN inference over device–edge–cloud further partitions operators to balance accuracy and throughput under network constraints [4].

### B. *Edge learning: federated, split and hybrids*

- **Federated Learning (FL).** Federated learning is a distributed training framework where each client (edge node) computes updates to a global model on its local data, and only those updates (e.g. gradient or weight differences) are sent to a central aggregator (or shared among peers) [25]. Crucially, the raw data never leaves the device – only learned parameters are exchanged – which preserves privacy. Federated learning addresses both data privacy and bandwidth issues – it drastically reduces communication compared to raw data transfer, as noted by McMahan et al. (2017) who achieved 10-100× fewer communication rounds than naive distributed SGD by using model averaging [25]. Recent surveys focus on efficient FL for foundation models, cataloguing parameter efficient finetuning (PEFT) in FL, communication/compute tradeoffs, and personalization under heterogeneity [26][27][28]. In civil aviation, research has indeed applied FL in scenarios like unmanned aerial vehicle networks and air traffic surveillance, demonstrating the approach's viability under aviation constraints [29].

- **Split learning (SL) and hybrids.** While federated learning shares parameters, split shares activation data to train a model jointly between a client and a server (or among multiple peers). By this coordinated forward-and-backward split, the full model is trained without any client exposing raw input data, while only intermediate features are communicated, which can be encrypted or obfuscated for privacy [30]. New work explores online split-point selection via multi-armed bandits to adapt to resource variation at the mobile edge [31]. Security analyses of SplitFed highlight poisoning and gradient-leakage risks, motivating encryption, secure aggregation and DP in safety-critical domains [32]. Split learning has various configurations (one-to-one, many-to-one, multi-hop) and can be combined with FL (as in SplitFed learning) for additional robustness [30]. For aviation, split learning offers a way to utilize powerful cloud resources for part of the model while still keeping raw data and partial computation at the edge, thereby blending the strengths of edge and cloud.

### C. *Edge intelligence benefits*

Fig. 1 provides an overview of the seven core benefits delivered by edge intelligence in safety - critical aviation environments, low task latency, enhanced privacy and security, computational efficiency, optimized bandwidth, offline capability, energy efficiency, and cost optimization.

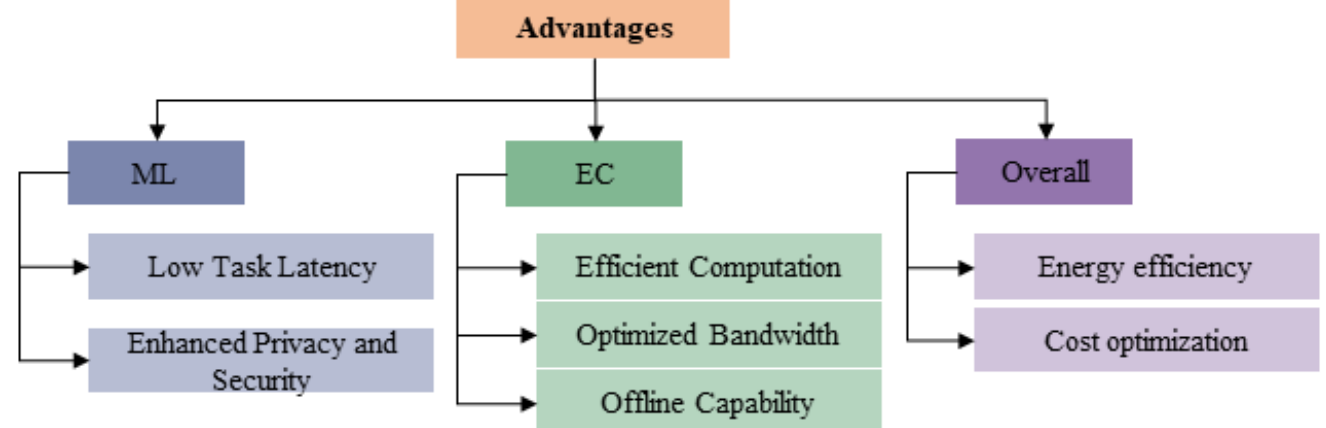

Fig. 1. Edge intelligence benefits

- **Low Task Latency.** By executing perception and decision layers on local devices or site-proximate edge servers, edge intelligence minimizes end-to-end response time from data capture to model inference and action and achieves sub-second reactions. or civil aviation, this latency headroom is essential for tower assistance, runway-incursion alerts, and flight-deck abnormal-event management where slower cloud round-trips would violate operational time budgets.

- **Enhanced Privacy and Security.** Edge intelligence enforces data minimization by transmitting features or decisions rather than raw streams, and reduces the attack surface by avoiding exposure over external backhaul links. For civil aviation, this directly supports regulatory compliance and protects operational know-how while enabling AI-assisted functions without widening the risk of data exfiltration.

- **Computational Efficiency.** Edge intelligence couples model compression with accelerator-aware operator scheduling and locality-aware memory access, it delivers captures higher throughput per watt and per byte of memory on resource-constrained platforms. For civil aviation, such efficiency allows advanced analytics to run reliably on certified on-board computers and ruggedized airport edge servers, reducing integration friction and easing long-term maintainability.

- **Optimized Bandwidth.** By filtering at the source, triggering on events, and summarizing into sparse detections or compressed evidence, edge intelligence uplinks only high-value information and smooths traffic, and thus reduces wide-area data volumes and their variability across constrained links. For civil aviation, preserving satellite or Aeronautical Telecommunication Network (ATN) capacity for safety-critical communications improves Quality-of-Service during peak operations and mitigates downstream storage burdens.

- **Offline Capability.** Offline capability ensures that AI services continue operating under degraded or absent connectivity. Since models and critical state reside locally, edge intelligence sustains functionality when ground links or satcom are unavailable. For civil aviation, this robustness is indispensable on oceanic and polar routes, at remote aerodromes, and during contingencies so that safety-relevant analytics never hinge on a live cloud connection.

- **Energy Efficiency.** By avoiding repeated long-haul transmissions, leveraging low-precision compute on accelerators, and enabling dynamic power management (e.g., early exit, duty-cycling), edge intelligence lowers the energy cost of intelligence at the source. For civil aviation, improved energy efficiency extends UAV endurance, reduces auxiliary power usage on aircraft and ground vehicles, and supports airport sustainability targets without compromising capability.
- **Cost Optimization.** Since edge intelligence reflects reduced total cost of ownership across compute, storage, and networking, it yields operational savings via faster turnarounds and fewer disruptions. For civil aviation, these savings improve margins in a cost-sensitive industry and make at-scale AI adoption feasible across fleets and airports.

## III. Organizational Computing Paradigms

Deploying edge intelligence in civil aviation involves architectural choices about how computing tasks are partitioned and orchestrated between devices, edge servers, and cloud data centers. We identify four paradigms of collaborative computing for edge AI, depicted in fig. 2, which differ in the roles played by cloud and edge resources and in the topology of coordination.Table I below summarizes paradigms of edge intelligence.

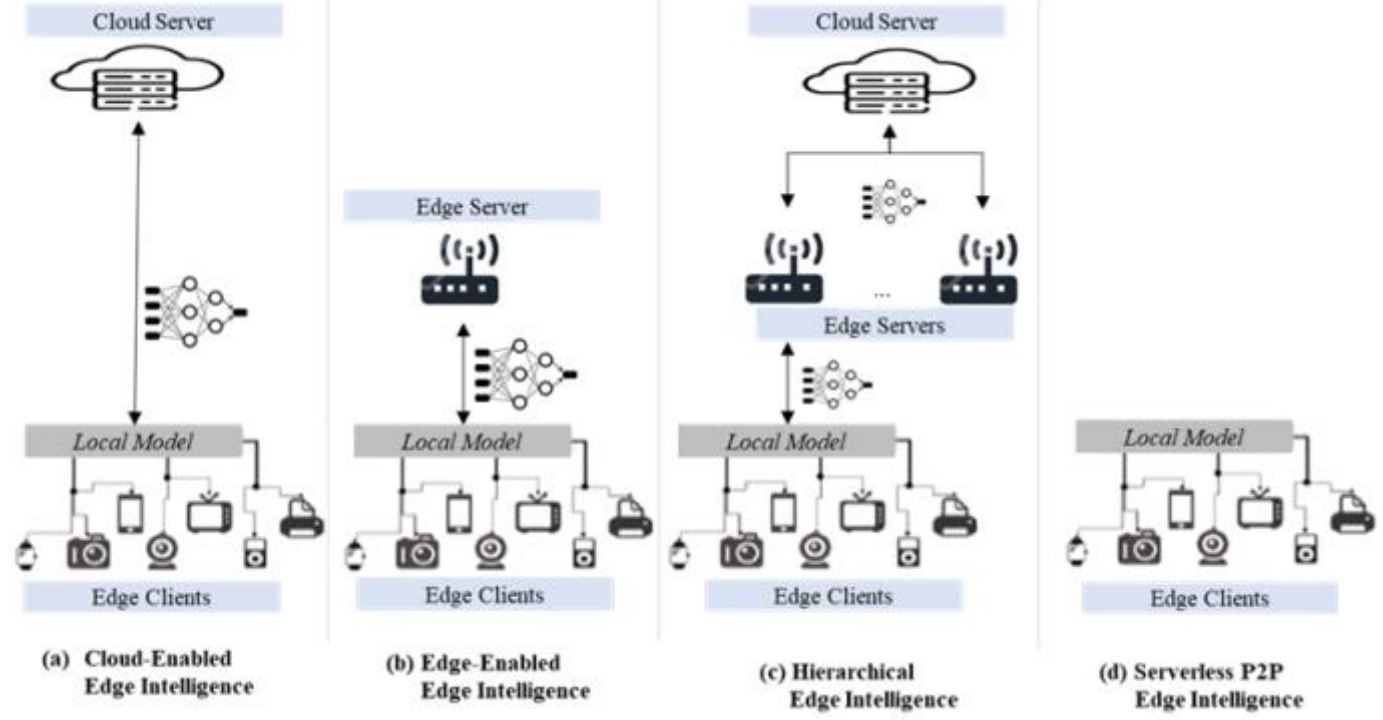

Fig. 2. Organizational computing paradigm of edge intelligence

TABLE I. Summary of Edge Intelligence Paradigms

| Paradigm | Network assumption | Data locality / privacy | Core strengths | Key limitations |
|---|---|---|---|---|
| Cloud-Enabled | Stable, adequate backhaul (e.g., gate Wi-Fi, fiber) | Raw/feature data may leave site; privacy review required | Access to heavy models; continuous retraining; global context | Connectivity dependence; mobility fragile; egress cost |
| Edge-Enabled | LAN-grade, low-latency, on-prem | Data stays on-site; strong sovereignty | Ultra-low latency; offline robustness; rapid iteration | Local capacity limits; per-site rollout/ops |
| Hierarchical | Mixed: real-time local + periodic sync upstream | Local processing with aggregated sharing; policy-controlled | Balanced resources; cross-site learning; failover | Orchestration complexity; policy/version management |
| Peer-to-Peer | Local ad-hoc links; intermittent tolerated | Fully local; no central data pool | No single point of failure; resilient; privacy-preserving | Consistency/consensus challenges; heterogeneous nodes; hard to audit |

- **Cloud-Enabled Edge Intelligence (Cloud–Device Collaboration).** In this paradigm, edge devices collaborate with a cloud server to perform AI inference or learning. Typically, the heavy computations are offloaded to the cloud, while the edge device does some light pre-processing or partial model execution. The cloud-enabled model places substantial reliance on the cloud's capabilities, effectively enabling edge devices to run AI that they otherwise could not.
- **Edge-Enabled Edge Intelligence (Edge–Device Collaboration):** In this paradigm, edge servers (or edge infrastructure) take on the role of the cloud by assisting end devices, with no involvement of a distant cloud. Essentially, intelligence is pushed further to the edge: a nearby server (e.g. at an airport data center, or a base station) collaborates with devices to perform inference/training. The devices offload tasks to this proximate edge, gaining low-latency access to greater compute than they possess, but all computation stays within the local network.
- **Hierarchical Edge–Cloud AI (Cloud–Edge–Device Collaboration):** This paradigm involves a multi-tier hierarchy of computing, where intelligence is distributed across cloud, edge, and device levels working in unison. Rather than choosing between cloud or edge, this model leverages both – assigning parts of the workload to devices, parts to nearby edge servers, and parts to the central cloud as appropriate. Hierarchical collaboration can be thought of as an extension of the first two paradigms combined: devices may offload some tasks to edge servers, which in turn might collaborate with the central cloud for additional processing or aggregated learning across sites. This paradigm is especially useful when tasks have components with different latency and locality needs.
- **Peer-to-Peer Edge AI (Serverless Device–Device Collaboration):** In this paradigm, edge devices collaborate directly with each other without any centralized server (neither cloud nor edge) coordinating them. It is a decentralized, serverless model of edge intelligence. Devices can share data or model updates among themselves in a peer-to-peer (P2P) network to collectively perform inference or training. This approach might involve gossip protocols, local ad-hoc networks, or blockchain-like ledgers to coordinate learning. Device–device collaborative inference, for instance, could split a neural network across multiple nearby devices that share intermediate results with each other to compute an output none could do alone.

## IV. Deployments And Applications Of Edge Intelligence In Civil Aviation

In this section, we translate the computing paradigms from section 3 into concrete deployment patterns for aircraft, towers, and airports, and then synthesize the applications from practice to show where edge intelligence is already delivering value in civil aviation.

### A. *From Paradigm to Deployment: Aviation-Specific Implementations*

Table II summarizes the potential implementations and deployment for edge intelligence in civil aviation.

TABLE II. Aviation-specific implementations for edge intelligence

| Paradigm | Cloud-Enabled | Edge-Enabled | Hierarchical | Peer-to-Peer |
|---|---|---|---|---|
| Typical deployment loci | Cameras, radios, kiosks (device) + regional/public cloud | Tower/airport MEC (Multi-access Edge Computing) rooms; on-aircraft edge server | Devices; airport edge; airline/private data center; regional cloud | UAV (Unmanned Aerial Vehicle) swarms; ad-hoc apron clusters; aircraft-to-aircraft |
| Operational latency target | 100 ms–seconds; lower than pure cloud, not hard-real-time | Sub-10–100 ms loops; site-critical | Real-time at device/edge; batch/global overnight | Very low local latency; soft real-time |
| Connectivity tolerance | Requires stable backhaul (gate Wi-Fi, fiber); tolerant to short gaps | Continues during WAN outage; relies on LAN/avionics bus | Mixed: real-time local; periodic northbound sync | Designed for intermittent links; infrastructure-light |
| Failover & resilience pattern | Graceful degradation to "edge-only pre-filters"; queue-then-process on reconnect | N+1 edge nodes; warm-standby VMs; local policy fallbacks | Edge sustains operations if cloud unavailable; cloud burst-assists during peaks | Gossip/consensus; role re-assignment on node loss |
| Update/MLOps cadence | Frequent cloud-side retraining; scheduled edge model refresh at shifts | Edge models patched during maintenance windows; periodic cloud pull for updates | Tiered: fast patching at edge; model-of-models retraining centrally | Opportunistic: on-mission exchange; post-mission consolidation |
| Integration complexity | Moderate: network + IDAM + data-sharing agreements | Site-by-site rollout; tight integration with CCTV/SMR/avionics | Higher: orchestration, lineage, policy/version control | Highest: consistency, security, auditability |
| Illustrative aviation fit | Airport video pre-filtering with cloud recognition; cloud ASR/NLU for Air Traffic Control (ATC) archives; connected aircraft maintenance post-flight | Runway/apron hazard detection; tower assistance; on-aircraft abnormal-event manager; passenger-services edge | Fleet health (Remaining Useful Life) and multi-airport ops; network-wide slot/crew optimization | Search-and-rescue mapping; collaborative multi-camera tracking; aircraft weather sharing during outages |

### B. *Representative Applications across the Aviation Value Chain*

In this section, we consolidate several representative applications in the real world and explain how their deployment choices directly follow the deployment from the last section.

- **ATC speech and tower assistance.** Large-scale ATC corpora (ATCO2) have enabled domain-adapted Automatic Speech Recognition (ASR) and Natural Language Understanding (NLU) for noisy Very High Frequency (VHF) communications, with 2024 releases detailing end-to-end pipelines from data capture to robust modeling [33][34]. Placing quantized or adapter-augmented ASR/NLU at towers and remote towers reduces dependence on wide-area backhaul and allows real-time alerts for call-sign confusion and read-back inconsistencies; in parallel, cloud components can continue improving language models between shifts, creating a practical edge-enabled plus cloud-refinement loop [33][34].

- **Airport surface safety and apron analytics.** Edge–cloud cooperative systems for apron surveillance (e.g., an "edge first, cloud verify" pattern) demonstrate that on-prem inference can flag candidate events quickly while keeping raw video on site, escalating only the most informative segments for heavier processing [35]. Follow-up studies localize non-cooperative abnormal behaviors from real apron datasets, indicating production-grade feasibility for ramp safety where latency and privacy must both be respected [36]. Together with cloud–edge orchestration patterns (frame filtering, model selection, cross-camera correlation), these results support Surface Movement Guidance and incursion prevention workflows in a repeatable manner [37][38].

- **Predictive maintenance and engine health management.** Recent work shows real-time Exhaust Gas Temperature (EGT) forecasting and Remaining Useful Life (RUL) estimation using deep learning, where on-prem edge screening provides rapid triage and defers deeper diagnostics to centralized systems when appropriate [39][40]. A systematic review consolidates data sources (Flight Data Recorder, maintenance logs) and modeling choices, and suggests architectures where federated or hierarchical learning enables fleet-level improvement without moving raw data off premises [41]. This "fast-edge + global-learn" pattern reduces mean time to action while preserving bandwidth and data sovereignty, aligning naturally with hierarchical deployments.

- **Operations sensing at non-towered and regional airports.** A computer-vision system deployed at non-towered airports demonstrates how local edge processing can produce structured operational data (e.g., movements, occupancy) with minimal backhaul, broadening situational awareness beyond major hubs and validating edge-only viability in lower-infrastructure environments [42].
- **UAV/UTM extensions and aerial edge computing.** Surveys on UAV-enabled edge computing and aerial Mobile Edge Computing (MEC) discuss resource-allocation and learning strategies under tight energy and bandwidth budgets, many of which transfer directly to aviation ground/air systems [43][44]. In practice, peer-to-peer or hierarchical variants can be tailored by mission: collaborative perception within a UAV swarm for emergency response, or regional edge nodes coordinating Unmanned Traffic Management (UTM) flows while cloud services synthesize longer-horizon traffic policies [43][44].

# V. Future Trends And Vision

Building upon the techniques and deployment patterns summarized earlier, edge intelligence in civil aviation is evolving along with broader technology trends, opening new trends and vision.

## A. *Edge-Capable Foundation Models and Multimodal Assistants*

A first trend is the emergence of edge-capable large multimodal models that deliver cockpit/tower assistants, on-premises summarization, and real-time perception without relying on wide-area backhaul. This vision hinges on a compression stack, i.e., post-training quantization, weight-only ultra-low precision, and adapter-based fine-tuning, that preserves accuracy while shrinking memory and compute footprints so models can execute on avionics-grade hardware or airport edge servers. Recent results on quantized fine-tuning and activation-aware weight quantization, as well as mixed dense–sparse schemes, show that instruction-tuned and multimodal models can retain quality at 8-bit, 4-bit, or even sub-4-bit precision with modest calibration effort [11][45][46]. Together, these techniques make it realistic to host procedure-aware copilots and tower briefers at the edge, with cloud used primarily for periodic model refreshes rather than in-the-loop inference.

## B. *Federated/Split Learning at Foundation-Model Scale*

A second trend is cross-operator collaboration without data sharing: airlines, airports, and OEMs collectively improve models while keeping raw telemetry, voice, and video in place. In practice this means federated learning (for privacy-preserving parameter aggregation) combined with split learning (for compute offload via feature-level collaboration), wrapped in parameter-efficient tuning to control bandwidth and on-device cost. Contemporary surveys and systems work detail how parameter-efficient fine-tuning (PEFT), personalization, and communication-efficient protocols extend to foundation models, and how adaptive split points mitigate heterogeneous edge resources [31][47]. For aviation, these hybrids enable fleet-wide Remaining Useful Life (RUL) predictors and airport-specific perception models to co-evolve without breaching data sovereignty.

## C. *Network-Native Orchestration across Device – Edge – Cloud*

A third trend is policy-driven placement of AI functions across devices, airport edges, and regional clouds, where the network acts as a scheduler: time-critical loops stay local; heavy analytics and global optimization shift to nearby or central compute when links permit. Cloud–edge cooperative video analytics and multi-agent orchestration frameworks already demonstrate end-to-end pipelines that filter at the source, escalate selectively, and learn centrally—patterns that map directly to apron safety, surface movement guidance, and tower assistance [48]. Over time, telemetry-aware policies will make this adaptive partitioning routine, improving latency–cost–accuracy trade-offs under varying load and connectivity.

## D. *Trustworthy and Certifiable Edge AI by Design*

A fourth trend is assurance-first engineering for edge AI in safety-critical operations: models calibrated under shift, outputs accompanied by evidence and uncertainty, and decision logic auditable end-to-end. Holistic evaluations and selective-trust methods provide the ingredients—benchmarks that expose multimodal failure modes and conformal alignment schemes that only surface predictions meeting statistical guarantees [45][49]. Embedding these ideas into edge pipelines (e.g., verify-then-say, reject options, provenance-linked replays) will accelerate regulatory acceptance and enable certification pathways for perception-and-policy stacks deployed on aircraft and at airports.

## E. *Robustness to Missing Modalities and Intermittent Links*

A fifth trend is graceful degradation when sensors fail, modalities are redacted, or connectivity drops. Two complementary lines of work are converging here: (i) pretraining and inference strategies that model modality reliability and reconstruct or re-weight from the context of available streams, and (ii) prompt-level correlation that conditions models to operate effectively when entire modalities are absent. Surveyed approaches to missing-modality learning and deep correlated prompting offer practical recipes for aviation scenarios where only speech+tracks or only video+telemetry are available at decision time [46][50]. The result is resilient situational awareness with explicit guarantees on performance under partial observability.

## F. *Synthetic Data and Digital-Twin Stress Testing*

A sixth trend is the use of controllable generators and high-fidelity digital twins to expand rare events (incursions, convective weather impacts, atypical failures), probe corner cases, and accelerate validation—without exposing real operational data. Configurable traffic-scenario generators and any-to-any multimodal synthesis models provide the building blocks to co-synthesize CVR/ATC audio, FDR/ADS-B tracks, video, and weather for targeted stress tests and curriculum design [51][52]. With domain-expert review and coverage metrics, these assets become a repeatable safety harness for pre-deployment training, adversarial red-teaming, and regression testing of edge stacks.

## VI. Conclusion

This paper delivers a unified technical background for edge intelligence in civil aviation by systematizing edge-inference and edge-learning techniques, mapping four collaborative computing paradigms to aviation environments, and synthesizing emerging deployments across aircraft, towers, and airports into actionable design guidance. Together, these elements provide a coherent reference architecture that links methods, infrastructure choices, and operational use cases.

Advancing the field now calls for certification-oriented evaluation (metrics, datasets, and assurance cases), privacy-preserving federation/splitting at foundation-model scale, robust device–edge–cloud orchestration under variable connectivity, and principled treatments of security, energy, and governance across stakeholders. Coordinated research and public–private pilots should stress-test edge stacks with digital-twin scenarios and establish repeatable MLOps for safety-critical operations. By providing a common vocabulary and design space, this paper aims to accelerate rigorous, low-latency, privacy-preserving AI that can be trusted at scale in the civil-aviation ecosystem.